\documentclass[10pt,twocolumn,letterpaper]{article}

\usepackage{cvpr}              
\usepackage[table]{xcolor}

\usepackage{graphicx}
\usepackage{algorithm}
\usepackage{algorithmic}
\usepackage{amsmath}
\usepackage{amsthm}

\newtheorem{theorem}{Theorem}[section]
\theoremstyle{definition}
\newtheorem{definition}[theorem]{Definition}

\usepackage{wrapfig}
\usepackage{appendix}
\usepackage{units}
\usepackage{xspace}

\definecolor{cvprblue}{rgb}{0.21,0.49,0.74}
\definecolor{purple}{rgb}{0.8,0.21,0.2}
\definecolor{yellow}{rgb}{0.8,0.8,0.2}
\definecolor{genericgreen}{rgb}{0,0.6,0}
\definecolor{personalizedblue}{rgb}{0,0.5,1}
\usepackage[pagebackref,breaklinks,colorlinks,citecolor=cvprblue]{hyperref}

\def\paperID{6019} 
\def\confName{CVPR}
\def\confYear{2024}

\title{FediOS: Decoupling Orthogonal Subspaces for Personalization\\ in Feature-skew Federated Learning}

\author{Lingzhi Gao$^{1*}$ \qquad Zexi Li$^1$\thanks{Equal contributions.} \qquad Yang Lu$^2$ \qquad Chao Wu$^{1}$\thanks{Corresponding author.}\\
$^1$Zhejiang University\qquad $^2$Xiamen University \\
{\tt\small \{lingzhigao,zexi.li,chao.wu\}@zju.edu.cn}\quad {\tt\small luyang@xmu.edu.cn}}

\begin{document}
\maketitle
\begin{abstract}
Personalized federated learning (pFL) enables collaborative training among multiple clients to enhance the capability of customized local models. In pFL, clients may have heterogeneous (also known as non-IID) data, which poses a key challenge in how to decouple the data knowledge into generic knowledge for global sharing and personalized knowledge for preserving local personalization. A typical way of pFL focuses on label distribution skew, and they adopt a decoupling scheme where the model is split into a common feature extractor and two prediction heads (generic and personalized). 
However, such a decoupling scheme cannot solve the essential problem of feature skew heterogeneity,
because a common feature extractor cannot decouple the generic and personalized features. 
Therefore, in this paper, we rethink the architecture decoupling design for feature-skew pFL and propose an effective pFL method called FediOS. 
In FediOS, we reformulate the decoupling into two feature extractors (generic and personalized) and one shared prediction head. Orthogonal projections are used for clients to map the generic features into one common subspace and scatter the personalized features into different subspaces to achieve decoupling for them.
In addition, a shared prediction head is trained to balance the importance of generic and personalized features during inference. Extensive experiments on four vision datasets demonstrate our method reaches state-of-the-art pFL performances under feature skew heterogeneity. 
\end{abstract}

\section{Introduction}
\label{sec:intro}

\begin{figure}[t]
    \centering
    \includegraphics[width=1.0\columnwidth]{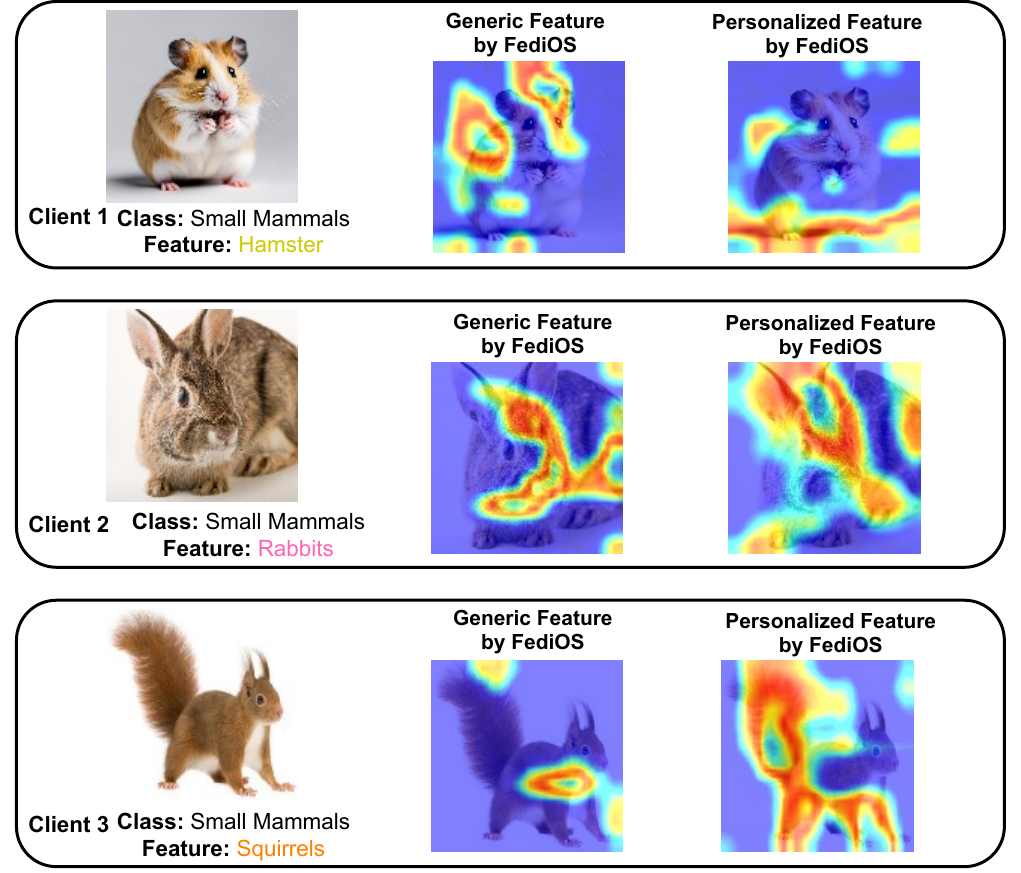}
    \caption{\textbf{Feature visualization of the proposed FediOS.} Three clients with the same class (small mammals) but different feature spaces (\textcolor{yellow}{hamster}, \textcolor{pink}{rabbits}, and \textcolor{orange}{squirrels}). \textit{It is found that FediOS can effectively decouple generic and personalized features.} According to small mammals, FediOS extracts the common \textbf{generic features} on the faces (mouths/eyes) and fur (color/texture). For the \textbf{personalized features}, they are decoupled as follows: \textcolor{yellow}{hamster}--the small and pink feet; \textcolor{pink}{rabbits}--the big ears; \textcolor{orange}{squirrels}--the furry tail and the relatively long feet.}
    \label{fig:fedios_feature_vis}
\end{figure}

Federated learning (FL)~\cite{FedAvg,canwe,Fedsurvey,PFLsurvey_new} is a distributed machine learning paradigm that enables collaborative training among multiple clients under the coordination of a central server. Unlike centralized learning, FL does not collect client data, therefore preserving data privacy and ownership~\cite{canwe,FedAvg}, also, it is communication-efficient~\cite{FedAvg,chen2021communication}. 

\begin{figure*}[h]
    \centering
    \includegraphics[width=1.8\columnwidth]{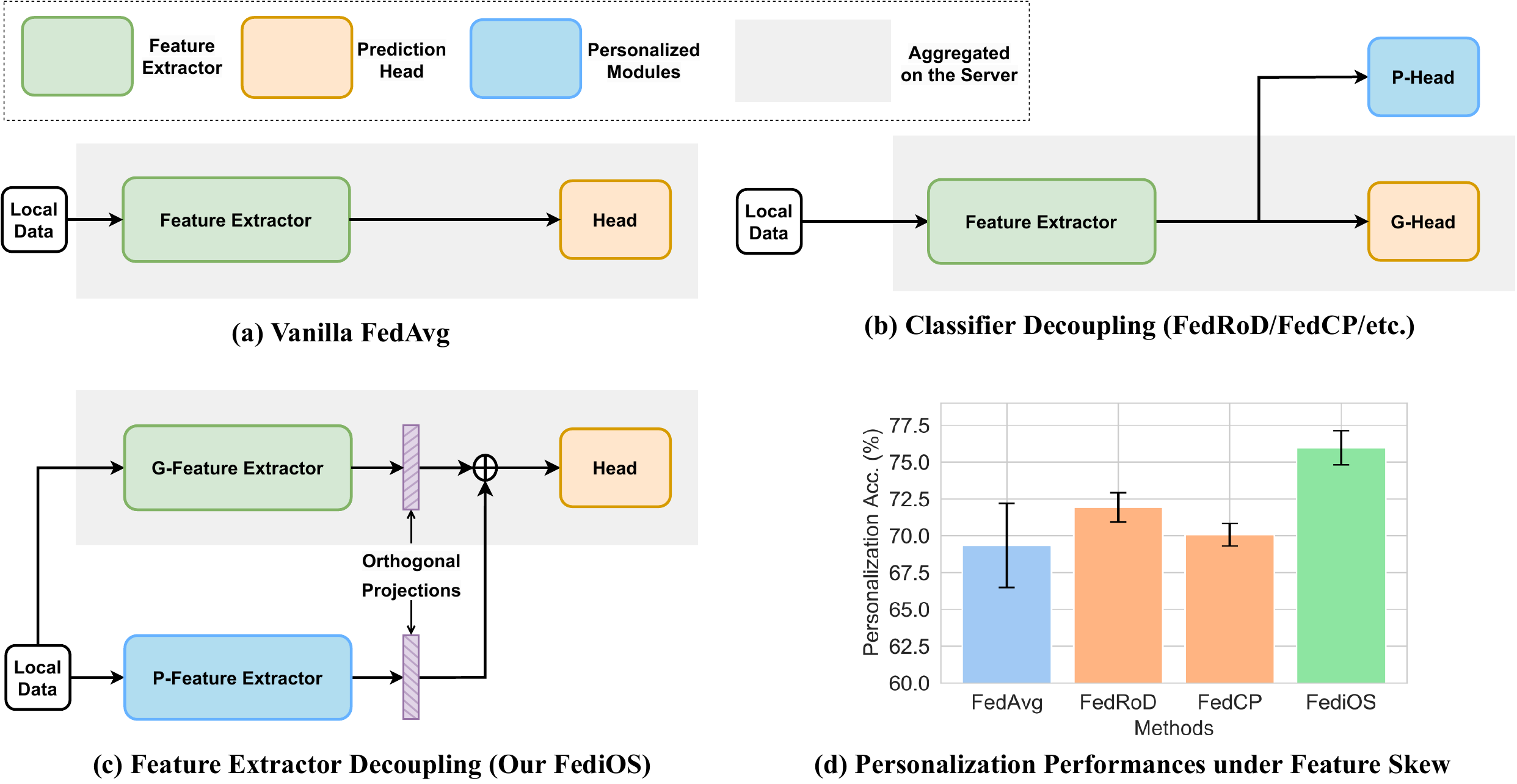}
    \caption{\textbf{Comparison of architecture decoupling designs.} \textbf{(a)} Vanilla FedAvg without decoupling. \textbf{(b)} Previous pFL methods decouple the prediction heads which may be ineffective under feature skew. \textbf{(c)} Our proposed FediOS decouples feature extractors through orthogonal subspaces. \textbf{(d)} Personalization performances under feature skew (Office-Caltech-10 with 4 clients). It is shown that prediction-head-decoupled methods have marginal improvements over FedAvg while our feature-extractor-decoupled FediOS have significant results.}
    \label{fig:fed_decouple_framework}
\end{figure*}

In FL, personalized federated learning (pFL) aims to improve the capability of clients' customized models through federated training~\cite{PFLsurvey,PFLsurvey_new,FedRoD,FedBN}. pFL addresses the concerns about why the clients need to join FL training (i.e., motivation) and what can they get from the FL training (i.e., incentive and credit). It has wide applications in computer vision, such as
image classification~\cite{FedBABU,FedRoD}, image segmentation~\cite{FedDP,dong2023federated}, and medical images~\cite{ chen2021personalized,zhang2023grace}. 
However, data heterogeneity is a key challenge in pFL, because different clients may have diverse preferences and user behaviors and, as a result, generate data that are non-IID with each other~\cite{FedNova,FedProx}. For learning and optimization, data heterogeneity means that clients have diverse learning objectives that are \textit{partially in common but partially inconsistent}. To realize better personalization, a fundamental research question is posted: 

\vspace{0.15cm}
\noindent\emph{How to decouple the data knowledge into generic knowledge for global sharing and personalized knowledge for preserving local personalization?}
\vspace{0.15cm}

To answer the question, we need to understand how data heterogeneity emerges in practice. 
Mainly, there are two types of data heterogeneity: \textbf{(a) label distribution skew}~\cite{zhang2022federated,FedRoD,Ditto,FedProx} which is described as the heterogeneity of $\mathcal{P}(y)$ among clients and \textbf{(b) feature skew}~\cite{pFedVEM,FPL_CVPR23,DFL} which refers to the heterogeneity of $\mathcal{P}(x)$ or $\mathcal{P}(x|y)$, where $x$ denotes the feature of data and $y$ is the label \footnote{Besides, in some less common cases such as clustered FL, heterogeneity of $\mathcal{P}(y|x)$ is also considered~\cite{CFL,IFCA,PANM}.}. Label distribution skew and feature skew are both common in practice. Take the photo classification as an example, some phone users may like dogs more than cats, so they have more dog photos (label distribution skew); for dog lovers, according to the same label of ``dog'', someone's dog is a Shiba while another user has a Husky, and different breeds of dogs have different visual features (feature skew).

Most previous works of pFL focus on label distribution skew~\cite{FedRep,FedRoD,FedCP,Ditto}. To decouple the generic and personalized knowledge, they adopt a decoupling scheme where the model is split into a common feature extractor and two prediction heads (generic and personalized)~\cite{FedRoD,FedCP,FedTHE,xu2022personalized} since the imbalance in labels will result in more drifts in the heads (e.g., for the classification task, the classifiers). The prediction-head-decoupled architecture is shown in Figure~\ref{fig:fed_decouple_framework} (b). However, for feature skew heterogeneity, one common feature extractor cannot suit all clients with divergent feature spaces, and decoupling the heads cannot address the problem since the generic and personalized features are encoded by the feature extractors, not the heads. 
In Figure~\ref{fig:fed_decouple_framework} (d), it is revealed that the methods of decoupling the heads have marginal improvements over vanilla FedAvg under feature skew. Therefore, rethinking the architecture decoupling design in feature-skew pFL is urgently needed.

In this paper, we reckon \textit{instead of decoupling the heads, we need to decouple the feature extractors for feature-skew pFL}. But how to effectively and efficiently decouple the features? Inspired by the orthogonal projections in continual learning, we propose \textbf{Fed}erated learning with decoupl\textbf{i}ng \textbf{O}rthogonal \textbf{S}ubspaces (\textbf{FediOS}). In FediOS, as illustrated in Figure~\ref{fig:fed_decouple_framework} (c), orthogonal projections are used for clients to map the generic features into one common subspace and scatter the personalized features into different subspaces. All clients share identical generic projections, whereas their personalized projections are mutually orthogonal and also orthogonal to the generic projections. We also devise a loss to regularize and strengthen the orthogonality. In addition, a shared prediction head is trained to balance the importance of generic and personalized features during inference. 

By leveraging the orthogonal subspaces, our FediOS can effectively decouple generic and personalized features, and an intuitive example is presented in Figure~\ref{fig:fedios_feature_vis}. In the figure, for the class of small mammals, FediOS extracts common features as generic features such as faces (mouths/eyes) and fur (color/texture), meanwhile, it also decouples the diverse personalized features such as furry tails for squirrels and big ears for rabbits. Additionally, from Figure~\ref{fig:fed_decouple_framework} (d), it is obvious that compared with previous decoupling schemes, FediOS can improve the personalization performances by a large margin under feature skew heterogeneity. We also note that the orthogonal projections in FediOS are set-and-fixed and learning-free, which is much more efficient than some feature disentanglement techniques that require training additional variational autoencoders (VAEs)~\cite{mathieu2019disentangling,ding2020guided}. 

To sum up, our contributions are as follows.
\begin{itemize}[leftmargin=*,nosep]
    \item We rethink the architecture decoupling design in feature-skew pFL and propose to decouple feature extractors instead of classifiers.
    \item We propose FediOS, which uses orthogonal projections to decouple generic and personalized features under the structure of dual feature extractors.
    \item Extensive experiments on four vision datasets, Office-Caltech-10, DomainNet, Digits-Five, and CIFAR-100, validate that our FediOS reaches state-of-the-art personalization performances under feature skew. Interestingly, it can also improve the generalization by effective generic knowledge sharing through decoupling.
\end{itemize}

\section{Related Work}
\label{sec:related}
\noindent\textbf{Federated Learning.} 
Since FedAvg~\cite{FedAvg} is first proposed, many federated learning methods are then designed to tackle data heterogeneity among clients for better generalization~\cite{FedProx,FedDyn}, personalization~\cite{PFLsurvey,PFLsurvey_new}, robustness~\cite{FedCorr, FedCNI, Ditto}, fairness~\cite{Ditto,zhang2021unified}, and efficiency~\cite{hamer2020fedboost, chen2021communication}. Besides personalization, the generalization of the global model is a key goal for FL training. It is deemed that better generalization of global models is the foundation of better personalization of local models~\cite{FedRoD,FedETF}. To achieve better generalization, algorithms tackling data heterogeneity from the perspectives of server-side aggregation and client-side local updates are both needed. For server aggregation, ensemble distillation~\cite{FedDF} and improved weighted aggregation~\cite{FedLAW,FedDisco} are proposed; while for client updates, techniques like proximal descent~\cite{FedProx}, dynamic regularization~\cite{FedDyn}, normalized gradients~\cite{FedNova}, and classifier calibration~\cite{luo2021no,FedETF} are tailored for resolving non-IID data, especially label distribution skew. For feature skew, a prototypical method based on prototype clustering is proposed to improve the generalization~\cite{FPL_CVPR23}.

\noindent\textbf{Personalized Federated Learning.} 
Personalized federated learning aims to improve the customized capability of clients' local models on the personalized local datasets~\cite{PFLsurvey,PFLsurvey_new}. Most of the previous works in pFL improve personalization under \textbf{label distribution skew}: Architecture decoupling is first proposed in FedRep~\cite{FedRep} to boost personalization. In FedRep, the feature extractor and the prediction head are decoupled and the server only aggregates the feature extractors while keeping the heads to be local and personalized. Afterward, FedRoD~\cite{FedRoD} further strengthens the decoupling design, as presented in Figure \ref{fig:fed_decouple_framework} (b), it is proposed to have two prediction heads, one for generalization (G-Head) with balanced empirical loss and another one for personalization (P-Head) with vanilla loss. FedRoD aggregates the feature extractor and the G-Head and keeps the P-Head local. Following the prediction-head-decoupled scheme in Figure \ref{fig:fed_decouple_framework} (b), FedTHE~\cite{FedTHE} balances the prediction weights of two heads for test-time robust personalization, and FedCP~\cite{FedCP} uses conditional policy to determine the proportion of global and local information within the two heads. Despite of its success in label skew, in this paper, we argue that the prediction-head-decoupled methods may not be effective enough under feature skew. Apart from decoupling, other methods are developed to improve personalization under label skew. FedBABU~\cite{FedBABU} proposes to fix the parameters of the head during the training process and only needs to fine-tune it locally at the end. FedNH~\cite{FedNH} improves personalization through prototypes, that it uses the smoothing-averaged prototypes as the classification heads. 

For \textbf{feature-skew personalization}, FedBN~\cite{FedBN} keeps the Batch Normalization layers of each client model locally and uploads other parts of the model for aggregation to solve the impact of domain shift between clients. 
pFedVEM~\cite{pFedVEM} is a framework that estimates uncertainty and model deviation among clients to derive a confidence value. This value then regulates the classifier head's weight during aggregation, facilitating the identification of common patterns and minimizing model discrepancies. In AlignFed~\cite{zhu2022aligning}, the model on each client is separated into a personalized feature extractor and a
shared classifier and collaboration is conducted by sharing and aggregating feature prototypes. By aggregating heads instead of feature extractors, it is the first attempt to rethink the decoupling under feature skew. But AlignFed does not go further into the decoupling that it cannot effectively decouple generic and personalized features by using one personalized feature extractor.
It is notable that our method has a novel decoupling scheme compared with previous algorithms and it is shown to be effective and efficient in decoupling generic and personalized features.

\section{Method}
\label{sec:method}

\subsection{Learning Objectives}
\noindent\textbf{Basic Settings.} We introduce a typical FL setting with $N$ clients holding potentially feature-skew heterogeneous data partitions $\mathcal D_{1}, \dots,\mathcal D_{N}$, respectively. For one data sample, we use $x$ to notate the data features and $y$ to denote the labels. The feature-skew heterogeneity among clients is defined as the following formulation: 
\begin{equation}
    \mathcal{P}_i(x|y) \neq \mathcal{P}_j(x|y), \forall i, j \in [N], i \neq j,
\end{equation}
where $\mathcal{P}_i(x|y)$ is the conditional probability of $x$ given $y$ for client $i$. The whole training data is the sum of all local data as $\mathcal D\triangleq \bigcup_{i}^{N} \mathcal D_i$. 

The model architecture $\Theta$ contains two parts, a feature extractor parameterized by $W$ and a prediction head parameterized by $H$, i.e., $\Theta = \{W, H\}$. The clients collaboratively train the models with data locally kept under the coordination of a central server. 

\noindent\textbf{Generalization Objective.} The global objective is to collaboratively train a generalized global model, which can be represented as:
\begin{align}
\min_{\Theta} \mathcal{L}(\Theta) = \sum_{i=1}^{N}  \frac{\lvert \mathcal D_{i} \rvert}{\lvert \mathcal D \rvert} \mathcal{L}_i(\Theta),
\label{equ:global_FL}
\end{align}
where $\lvert \mathcal D \rvert$ is the number of data samples and $\mathcal{L}_i(\Theta)$ denotes the empirical risk of client $i$ as following:
\begin{align}
\mathcal{L}_i(\Theta) = \sum_{k=1}^{\lvert \mathcal D_{i} \rvert}  \frac{1}{\lvert \mathcal D_{i} \rvert} l({y}_{k},f(x_k;\Theta)),
\label{equ:global_FL2}
\end{align}
where $(x_k,y_k)\in \mathcal D_{i}$ is the data of client $i$, $f(\cdot;\cdot)$ is the model function that generates outputs given the parameters, and $l(\cdot,\cdot)$ is the loss function that measures the difference between the outputs of the model and the truth labels.

\noindent\textbf{Personalization Objective.} In pFL, every client aims to have their own personalized models that can perform better personalization on their own datasets. The learning objective of pFL is:
\begin{align}
\min_{\{\Theta_1,\dots ,\Theta_N\} \in \mathbf{\Theta}} \mathcal{L}(\mathbf{\Theta}) = \sum_{i=1}^{N}  \frac{\lvert \mathcal D_{i} \rvert}{\lvert \mathcal D \rvert} \mathcal{L}_i(\Theta_i),
\label{equ:PFL}
\end{align}
where $\Theta_i$ denotes the personalized model parameters for client $i$. Finally, we will get a set of personalized models $ \mathbf{\Theta}^{\star}=\{\Theta_1^{\star},\dots,\Theta_N^{\star}\}$ to better adapt to clients' local data distributions.

We note that the generalization and the personalization objectives are not conflicted. As discussed in the literature, great generalization is the foundation of great personalization performances~\cite{FedRoD,FedETF}. Also, in this paper, we discover how to decouple the generic/generalized representation knowledge from clients to enhance local personalization better under feature-skew heterogeneity.

\subsection{Decoupling Orthogonal Feature Subspaces}

Instead of decoupling the heads, for feature-skew heterogeneity, we propose to decouple the feature extractor into two, one for generic feature extraction and another for personalized features. \textit{But how to realize such an effective decoupling?} A straightforward way is resorting to feature disentanglement techniques that are commonly used in feature debiasing~\cite{mathieu2019disentangling,ding2020guided}. However, disentanglement techniques usually require training an additional VAE, which is computation-expensive at edge clients. Thus, to enable efficient and effective decoupling, we adopt the orthogonal subspace technique, which was previously used in continual learning for solving the catastrophic forgetting problem~\cite{CL}. Compared with the VAE methods, the orthogonal subspace method\cite{CL} is set-and-fixed and learning-free, and it is highly effective in solving our problem. 
We first give the definition of orthogonal subspace as follows.

\begin{definition}[Orthogonal Subspace]
	\label{def:ospace}
The vector space $S$ has two subspaces $S_1$ and $S_2$. $S_1$ and $S_2$ are said to be orthogonal if any vectors in each subspace are pair-wisely orthogonal, written as:
\begin{align}
\mathbf{v_1} \cdot \mathbf{v_2} = 0, \quad \forall \mathbf{v_1} \in S_1, \mathbf{v_2} \in S_2.
\label{equ:ospace}
\end{align}
\end{definition}

\noindent\textbf{Orthogonal projections facilitate orthogonal feature subspaces.} 
In continual learning, orthogonal projections are used to enable the time-varying data to be learned into different subspaces of one model~\cite{CL}. Inspired by this, in our work, we use orthogonal projections to learn orthogonal feature representations of two different feature extractors, i.e., the generic and personalized. 

We use the projection matrices that satisfy the following orthogonal and normalized properties:
\begin{align} \label{equ:p_matrix}
{P}_{i}^\top{P}_{i} = I;\:
{P}_{i}^\top{P}_{j} = 0, i \neq j.
\end{align}

Given the raw data feature $x$, we assume the feature extractor $W$ produces the extracted feature as $\phi_W = f(x; W)$. 
After applying a projection matrix $P_i$, the extracted feature is $\phi_{P_i} = P_i \phi_W$. We will show that by applying orthogonal projections $P_i$ and $P_j$ that satisfy Equation (\ref{equ:p_matrix}), we can have orthogonal gradient updates, as a result, obtaining the orthogonal model parameters and feature spaces.

Assuming that each feature extractor consists of $L$ hidden layers, an input sample $x$ will be passed in each layer as follows: the $l$-th layer's output is $h_l = \sigma(h_{l-1}; W_l)$, where $\sigma(\cdot)$ is an activation function, $W_l$ is the parameter of the $l$-th layer, $h_0 = x$, and $h_L = \phi_W$. Therefore, $\phi_{P_i} = P_i h_L$. In the training process, for any projection matrix $P_i$, the gradient of any intermediate layer $h_l$ can be decomposed into:
\begin{align}
\notag g_l^{P_i} = \frac{\partial l}{\partial h_l} = \left( \frac{\partial l}{\partial h_L} \right)\frac{\partial h_L}{\partial h_l} = \left( \frac{\partial \phi_{P_i}}{\partial h_L}\frac{\partial l}{\partial \phi_{P_i}} \right)\frac{\partial h_L}{\partial h_l} \\
 = \left( P_i\frac{\partial l}{\partial \phi_{P_i}} \right)\prod_{k=l}^{L-1}\frac{\partial h_{k+1}}{\partial h_k} = g_L^{P_i}\prod_{k=l}^{L-1}D_{k+1}W_{k+1},
\end{align}

where $D_{k+1}$ is the Jacobian matrix and we assume it to be an identity. It can be seen that for any layer in the feature extractor, its gradient is associated with the last layer's gradient $g_L^{P_i}$; and for orthogonal projections $P_i \perp P_j$, the last layer's gradients are orthogonal that $g_L^{P_i} \perp g_L^{P_j}$. Due to the orthogonal gradient updates, we can have orthogonal model parameters and feature spaces.

\begin{figure}[t]
    \centering
    \includegraphics[width=1.1\columnwidth]{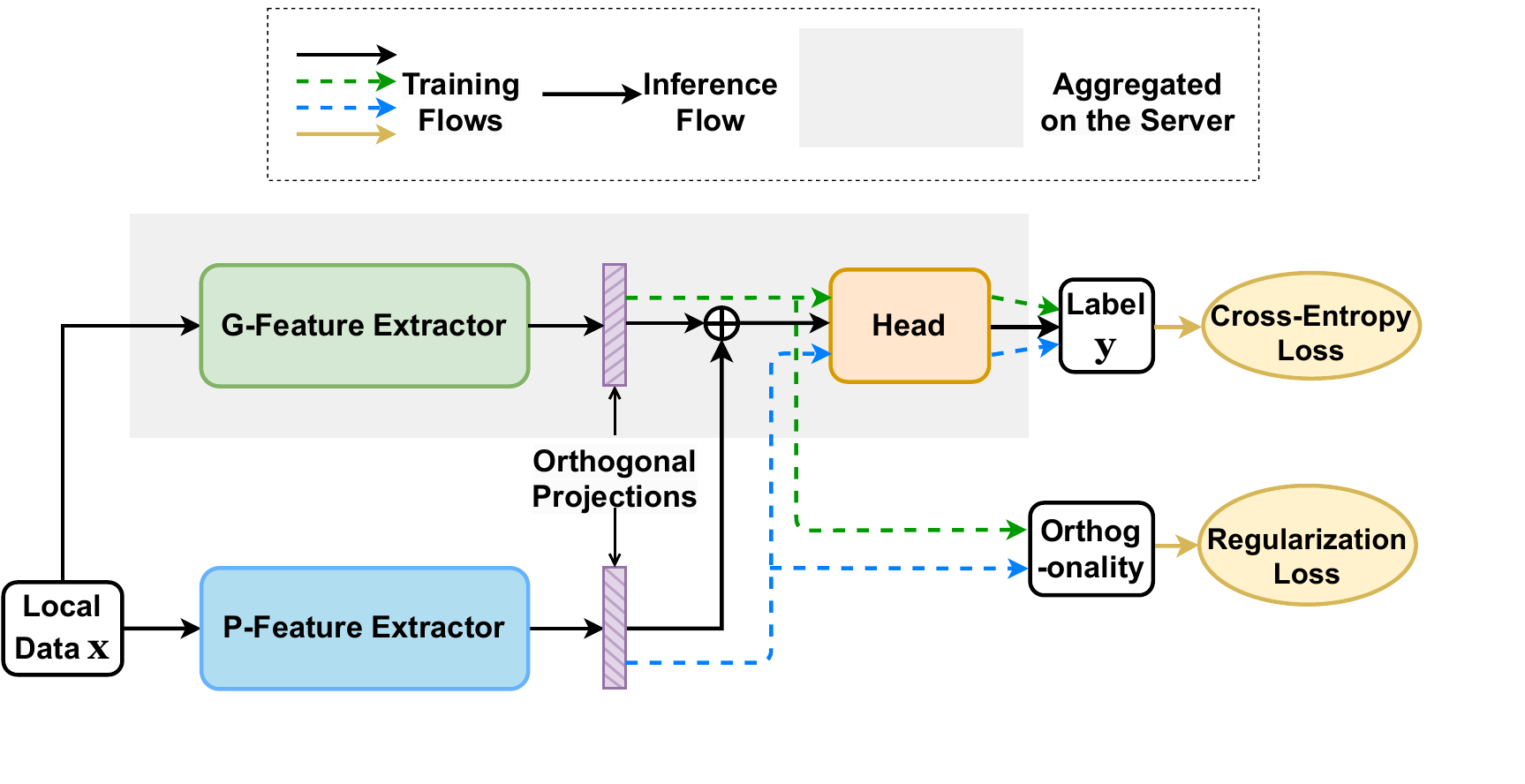}
    \vspace{-0.9cm}
    \caption{\textbf{Training and inference illustrations of FediOS.}}
    \label{fig:fedios_framework}
\end{figure}

As shown in Figure~\ref{fig:fedios_framework}, for each client $i \in [N]$, we have a generic feature extractor (parameterized by $W^g_i$), a personalized feature extractor (parameterized by $W^p_i$) with the same structure, and one prediction head (parameterized by $H_i$). All clients have the same generic projection $P^g$ for the generic feature extractor and have distinct personalized projections $P^p_i,\ \forall i \in [N]$ for their personalized extractors. All the projections $\mathbf{P} = \{P^g, P^p_{1},\dots, P^p_{N}\}$, including one generic and $N$ personalized, are synthesized at initialization and fixed during training, and they satisfy the orthogonal and normalized properties in Equation (\ref{equ:p_matrix}) pair-wisely. This ensures that all the generic feature extractors are learned in a common subspace and all the personalized extractors are orthogonal with each other and are also orthogonal with the generic extractor.

\subsection{Training Stage}
For each client $i, \ \forall i \in [N]$, given a data sample $(x_k, y_k) \in \mathcal D_{i}$, the generic feature $\phi_{i,k}^{g}$ and the personalized feature $\phi_{i,k}^{p}$ are given by:
\begin{align}
\phi_{i,k}^{g} = P^g f({x}_k;W^g_i),\:
\phi_{i,k}^{p} = P^p_i f({x}_k;W^p_i).
\label{equ:pFea}
\end{align}

We also have an fused feature representation $\phi_{i,k}^{f}$:
\begin{align}
\phi_{i,k}^f = \alpha\phi_{i,k}^{g} + (1-\alpha)\phi_{i,k}^{p},
\label{equ:Fea}
\end{align}
where $\alpha$ is the hyperparameter controlling the proportion of the generic feature. Generally, $\alpha$ is set around 0.5 to have a balance between the generic and personalized features. Interestingly, we will show in the experimental part (Figure~\ref{fig:alpha_acc_curves}), that different datasets may have slightly different optimal $\alpha$, indicating the intrinsic generic-or-personalized feature structures of the datasets. 

\noindent\textbf{The Cross-Entropy Loss.} 
During the training process, we employ the cross-entropy loss function to optimize the model parameters. By using one common prediction head ${H}_i$, we compute the cross-entropy loss for the three features $\phi_{i,k}^{f},\phi_{i,k}^{g}, \phi_{i,k}^{p}$, respectively. 
\begin{align}
\notag \mathcal {L}_{ce} = &-\frac{1}{\lvert \mathcal D_{i} \rvert}\bigg( \underbrace{\sum_{k=1}^{\lvert \mathcal D_{i} \rvert} {y}_{k} \log (f({\phi_{i,k}^f};{H}_i))}_{\text{fused features}} + \\ 
 &\underbrace{\sum_{k=1}^{\lvert \mathcal D_{i} \rvert} {y}_{k} \log (f({\phi_{i,k}^{g}};{H}_i))}_{\text{\textcolor{genericgreen}{generic features}}}+ \underbrace{\sum_{k=1}^{\lvert \mathcal D_{i} \rvert} {y}_{k} \log (f({\phi_{i,k}^{p}};{H}_i))}_{\text{\textcolor{personalizedblue}{personalized features}}}\bigg).
\label{equ:lossce}
\end{align}
In this way, the prediction head is learned to balance the tradeoff between the generic and the personalized features, and the two feature extractors are both trained to effectively contribute to the personalization task.

\noindent\textbf{The Orthogonality Regularization Loss.}
Additionally, we utilize the absolute value of the inner product between generic and personalized features as a loss function to strengthen the orthogonality between these two features:
\begin{align}
\mathcal {L}_{re} = \frac{1}{\lvert \mathcal D_{i} \rvert}\sum_{k=1}^{\lvert \mathcal D_{i} \rvert}{\lvert \phi_{i,k}^{g} \cdot \phi_{i,k}^{p} \rvert}.
\end{align}

Finally, the overall loss function is:
\begin{align}
\mathcal{L} = \mathcal {L}_{ce} + \lambda\mathcal {L}_{re},
\label{equ:loss}
\end{align}
where $\lambda$ is the hyper-parameter to control the strength of the regularization loss. The clients conduct Equation (\ref{equ:loss}) for $E$ epochs and communicate with the server.

\noindent\textbf{Server-side Aggregation.}
After the client $i$ completes local training, the generic feature extractor $W^g_i$ and the prediction head $H_i$ will be uploaded to the server and aggregated by FedAvg~\cite{FedAvg}. The overall number of communication and local training rounds is $T$. 
The training procedure is shown in Figure~\ref{fig:fedios_framework} and Algorithm~1 in the appendix.

\subsection{Inference Stage}
During the personalization inference stage on the client side, we only use the fused features defined in Equation (\ref{equ:Fea}), and the learned prediction head can implicitly balance the tradeoff between the generic and personalized features. The inference workflow is also illustrated in Figure~\ref{fig:fedios_framework}. To validate the global model on the server, we use only the aggregated generic feature extractor, the generic projection, and the aggregated prediction head.

\section{Experiment}
\label{sec:experiment}

\begin{table*}[!ht]
    \footnotesize
    \centering
      \vspace{-0.5cm}
    \caption{\textbf{Results in terms of personalization accuracy (\%) of local models on four datasets under different numbers of clients ($N$).} The best methods in each setting are highlighted in \textbf{bold} fonts.}
    \begin{tabular}{l|cc|cc|cc|cc}
    \toprule
    Dataset&\multicolumn{2}{c}{Office-Caltech-10}&\multicolumn{2}{c}{DomainNet}&\multicolumn{2}{c}{Digits-Five}&\multicolumn{2}{c}{CIFAR-100}\\
    \cmidrule(lr){1-3}
    \cmidrule(lr){4-5}
    \cmidrule(lr){6-7}
    \cmidrule(lr){8-9}
    Number of Clients ($N$) &\multicolumn{1}{c}{4}&\multicolumn{1}{c}{12}&\multicolumn{1}{c}{6}&\multicolumn{1}{c}{12}&\multicolumn{1}{c}{5}&\multicolumn{1}{c}{10}&\multicolumn{1}{c}{5}&\multicolumn{1}{c}{10}\\
    \midrule

     Local &70.67{±1.56}	&46.71{±1.31}	
   &38.34{±1.69}	&30.45{±0.82}	&85.41{±0.57}	&79.17{±0.24} &62.06{±0.74}	&49.20{±4.71}\\
    \midrule
    FedAvg$_{\textcolor{personalizedblue}{2017}}$ \cite{FedAvg}	&69.34{±2.86}	&55.47{±1.54} &36.40{±3.35}	&29.50{±1.22}	&86.78{±0.74}	&86.46{±0.72} &54.87{±1.58} &52.81{±0.13}	\\
    \midrule
    FedDyn$_{\textcolor{personalizedblue}{2021}}$ \cite{FedDyn}&69.57{±1.96}	&56.03{±0.25}	&39.47{±3.23}	&26.98{±2.39} &84.80{±0.14}	&84.97{±0.35} &45.45{±0.41}	&43.07{±1.76} \\
    MOON$_{\textcolor{personalizedblue}{2021}}$ \cite{MOON}&63.15{±6.56}	&52.74{±5.60}	&35.02{±2.89}	&27.45{±0.64} &84.16{±0.11}	&78.92{±0.97} &52.40{±0.37}	&50.20{±0.41} \\
    \midrule
    Ditto$_{\textcolor{personalizedblue}{2021}}$ \cite{Ditto} &72.04{±0.40}	&53.63{±0.59}	&43.13{±0.89}	&33.83{±0.50} &81.77{±0.38}&72.98{±0.16} &58.25{±0.61}	&45.88{±4.27}    \\
    FedProto$_{\textcolor{personalizedblue}{2022}}$ \cite{FedProto}&70.32{±1.53}	&49.25{±1.07}	&40.17{±1.23}	&31.04{±0.65}	&87.23{±0.28}	&82.41{±0.05} &60.27{±0.40}	&50.01{±3.85}\\
    \midrule
    FedRoD$_{\textcolor{personalizedblue}{2022}}$ \cite{FedRoD}&71.94{±0.99}	&57.45{±2.04}	&40.50{±3.06}	&32.40{±3.19} &86.39{±0.23}	&86.72{±0.12} &58.87{±0.76}	&56.27{±0.82}\\
    FedCP$_{\textcolor{personalizedblue}{2023}}$  \cite{FedCP}&70.07{±0.77} &52.16{±0.78} &41.36{±1.19} &29.51{±1.38} &87.28{±0.09}	&86.48{±0.09} &54.21{±0.46} &53.96{±0.72}	\\
    \midrule
    FedBN$_{\textcolor{personalizedblue}{2021}}$ \cite{FedBN}&71.75{±1.31}	&59.70{±0.22}	&41.64{±3.41}	&36.79{±1.02}	&87.17{±0.54}	&86.82{±0.57} &54.87{±0.58}	&52.81{±0.13}\\
    pFedVEM$_{\textcolor{personalizedblue}{2023}}$ \cite{pFedVEM}&70.32{±2.58} &58.86{±2.47} &39.85{±2.50} &27.56{±1.19}&81.28{±1.05}	&80.11{±0.71} &55.08{±0.50} &51.15{±1.69}\\
    \midrule
    \rowcolor{gray!20}\textbf{Our FediOS} &\textbf{75.98{±1.16}} &\textbf{60.60{±0.80}}  &\textbf{45.90{±0.37}}  &\textbf{37.32{±0.43}} &\textbf{88.65{±0.49}}	&\textbf{86.94{±0.10}} &\textbf{64.99{±0.72}} &\textbf{58.00{±2.17}} \\
    \bottomrule
    \end{tabular}
    \label{table:sota_table}
\end{table*}

\subsection{Experimental Settings}
\textbf{Datasets and Models.} Following previous works~\cite{FedBN,pFedVEM,zhu2022aligning,DFL}, we use four vision datasets to simulate feature-skew heterogeneity in FL: Office-Caltech-10 \cite{office}, DomainNet \cite{domainnet}, Digits-Five \cite{domainnet}, and CIFAR-100 \cite{cifar100}. \textit{\textbf{(1) Feature skew with domain shifts.}} Office-Caltech-10 (4 domains), DomainNet (6 domains), and Digits-Five (5 domains) are the datasets that contain domain shifts. By using these datasets in FL, clients have heterogeneous datasets with feature-skew domain shifts~\cite{zhu2022aligning,FedBN}. 
\textit{\textbf{(2) Natural feature skew with different subclasses.}} CIFAR-100 contains 20 superclasses and each superclass contains 5 subclasses. It is found that within each superclass, images from different subclasses have natural feature skew, where common feature patterns and distinct/personalized features may both exist. Following previous work~\cite{pFedVEM}, we use the superclasses as the prediction labels and craft natural feature skew by assigning different subclasses to clients. For the models, if not mentioned otherwise, we use ResNet18 \cite{ResNet18} for Office-Caltech-10 and DomainNet, a six-layer CNN \cite{FedBN} for Digits-Five, and a four-layer CNN \cite{FedCP} for CIFAR-100. Due to space limits, for more data and model details, please refer to the appendix.


\noindent\textbf{Compared Methods.} We use 5 sets of 9 FL algorithms as baselines that are strong or are most relevant to our method. 
\textit{\textbf{(1) Vanilla FedAvg.}} 
FedAvg~\cite{FedAvg} is the first FL algorithm but is also shown to be a strong baseline in pFL scenarios~\cite{FedRoD}. 
\textit{\textbf{(2) Generalized FL methods.}} 
Generalization is the foundation of personalization, therefore, we compare the state-of-the-art FL methods targeting generalization and see how they perform under feature-skew personalization. It includes FedDyn~\cite{FedDyn}, FL with dynamic regularization; MOON~\cite{MOON}, FL with model contrastive learning. 
\textit{\textbf{(3) General pFL methods.}} 
We compare two common and strong pFL methods, which are Ditto~\cite{Ditto}, personalization by using global models to regularize local model training; FedProto~\cite{FedProto}, personalization by using prototypes to extract information from a class-based perspective. 
\textit{\textbf{(4) Prediction-head-decoupled pFL methods.}} 
We compare the pFL methods decoupling the prediction heads, which are shown to be state-of-the-art in label distribution skew~\cite{FedRoD,FedCP}. FedRoD~\cite{FedRoD}, pFL method using balanced and personalized losses respectively in two heads. FedCP~\cite{FedCP}, separating feature information via conditional policy in two heads. 
\textit{\textbf{(5) pFL methods tailored for feature skew.}} 
We also validate previous feature-skew pFL methods. 
FedBN~\cite{FedBN}, personalization by retaining the Batch Normalization layer. pFedVEM~\cite{pFedVEM}, the state-of-the-art feature-skew pFL method with adjusting aggregation via variational expectation maximization. \looseness=-1

\noindent\textbf{Client Settings.} 
We set the number of clients $N$ to be a multiple of the number of domains or superclasses, and each client only has one domain/superclass's data. Specifically, $N \in \{4, 12\}$ for Office-Caltech-10, $N \in \{6, 12\}$ for DomainNet, $N \in \{5, 10\}$ for Digits-Five, and $N \in \{5, 10\}$ for CIFAR-100. In each round, we test the local models after training~\cite{FedRoD} on clients' local test datasets and take the average accuracy of clients as the personalization performance~\cite{FedRoD,FedCP}. We record the personalized performance in each round and take the average of the last 5 rounds as the result. The settings of $E$ and $T$ are in the appendix.

\noindent\textbf{Implementation.} In all experiments, we conduct experiments under three different seeds and present the mean accuracy and the standard deviation.

\begin{figure*}[t]
 \vspace{-0.5cm}
\centering
    \begin{subfigure}[b]{0.245\textwidth}
        \centering
        \includegraphics[width=\linewidth]{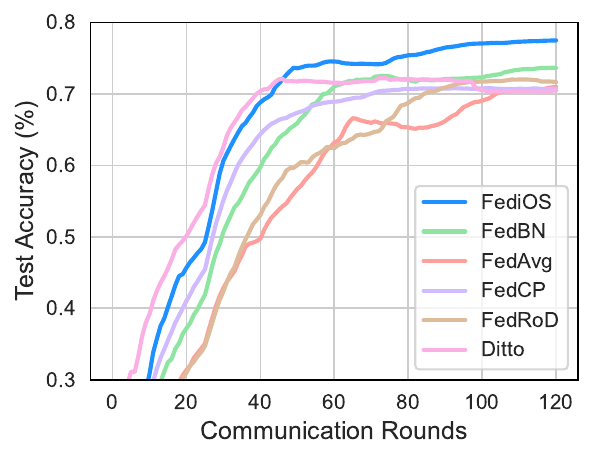}
        \caption{Office-Caltech-10 with $N=4$.}
    \end{subfigure}
    \begin{subfigure}[b]{0.245\textwidth}
        \centering
        \includegraphics[width=\linewidth]{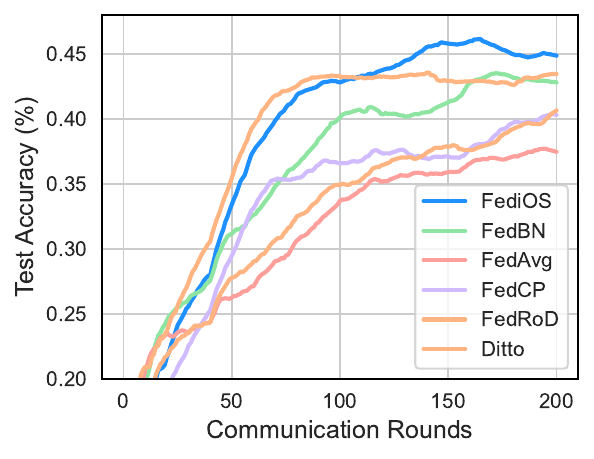}
        \caption{DomainNet with $N=6$.}
    \end{subfigure}
    \begin{subfigure}[b]{0.245\textwidth}
        \centering
        \includegraphics[width=\linewidth]{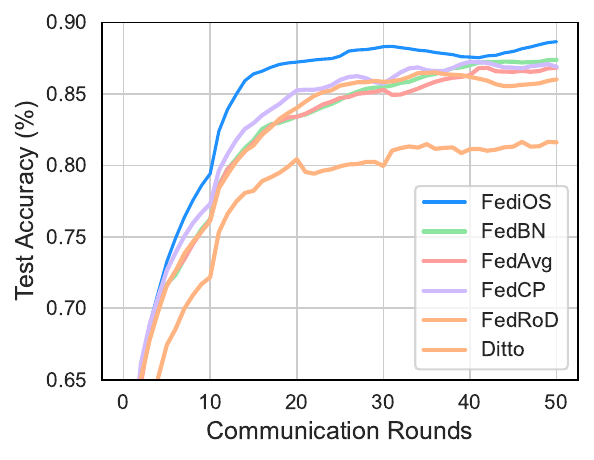}
        \caption{Digits-Five with $N=5$.}
    \end{subfigure}
    \begin{subfigure}[b]{0.245\textwidth}
        \centering
        \includegraphics[width=\linewidth]{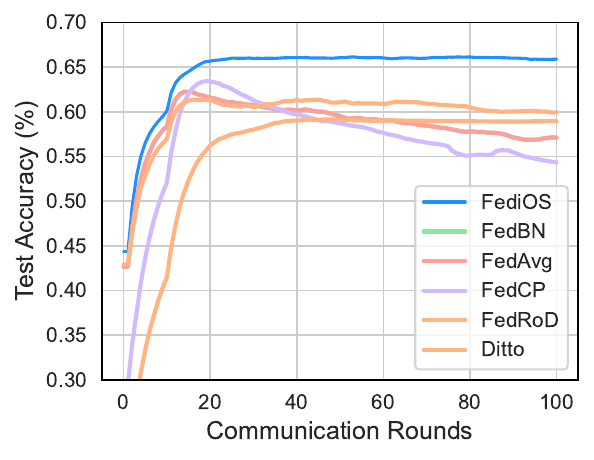}
        \caption{CIFAR-100 with $N=5$.}
    \end{subfigure}
    \vspace{-0.3cm}
\caption{\textbf{Personalization test accuracy curves of the methods under various datasets.} 
}
\label{fig:acc_curves}
\vspace{-0.2cm}
\end{figure*}

\begin{table*}[!h] 
	\centering
\begin{minipage}{0.475\linewidth}
    \centering
	\tiny
	\caption{\small \textbf{Results in terms of  generalization accuracy (\%) of global models.}} 
    \vspace{-2em}
    \resizebox{\linewidth}{!}{
	\begin{tabular}{lc|c|c|c}
	\toprule
    Dataset & \multicolumn{2}{c}{Digits-Five} & \multicolumn{2}{c}{CIFAR-100}  \\
       \cmidrule(lr){1-1}
    \cmidrule(lr){2-3}
    \cmidrule(lr){4-5}
    Number of Clients & 5 & 10 & 5 & 10\\
    \midrule
    FedAvg \cite{FedAvg} &83.89{±0.51} &83.45{±0.29} &49.00{±0.20} &48.38{±0.74}  \\ 
    \midrule
    Ditto \cite{Ditto}  &84.15{±0.27} &83.45{±0.18} &48.42{±0.06} &48.65{±0.73}  \\ 
    FedRoD \cite{FedRoD} &83.80{±0.10} &83.41{±0.18} &48.55{±0.52} &48.44{±0.43} \\
    FedCP \cite{FedCP} &83.87{±0.13} &82.86{±0.27} &41.63{±0.52} &42.79{±0.31} \\
    pFedVEM \cite{pFedVEM} &77.94{±1.02} &78.52{±0.34}&42.73{±0.66} &44.16{±0.46} \\
    \midrule
   \rowcolor{gray!20}\textbf{Our FediOS} &\textbf{84.47{±0.19}} &\textbf{83.85{±0.25}} &\textbf{49.05{±0.24}} &\textbf{48.83{±0.30}} \\
    \bottomrule
	\end{tabular}}
	\label{table:global_results}
    \end{minipage}
     \hfill
	\begin{minipage}{0.51\linewidth}\centering
	\tiny
    \caption{\small \textbf{Results (\%) under various numbers of clients with partial client sampling.} The dataset is CIFAR-100.}
    \resizebox{\linewidth}{!}{
    \begin{tabular}{l|c|c|c|c}
    \toprule
    Number of Clients&\multicolumn{2}{c}{15} &\multicolumn{2}{c}{20}\\
    \cmidrule(lr){1-1}
    \cmidrule(lr){2-3}
    \cmidrule(lr){4-5}
    Methods/Sampling rate &0.6 &1.0 &0.6 &1.0\\
    \midrule
    FedAvg \cite{FedAvg} &52.18{±0.51} &51.75{±0.61}&45.40{±1.53} &46.80{±1.00}\\
    FedCP \cite{FedCP} &54.22{±0.17} &53.14{±0.37} &47.54{±2.84} &47.60{±1.99}\\
    FedRoD \cite{FedRoD}  &54.51{±0.06} &54.82{±0.65} &49.89{±0.87} &49.40{±1.50}\\
    pFedVEM \cite{pFedVEM} &49.32{±0.74} &49.80{±0.04} &48.64{±1.21} &48.45{±0.82}\\
    \midrule
   \rowcolor{gray!20}\textbf{Our FediOS} &\textbf{54.93{±0.31}} &\textbf{55.59{±0.37}} &\textbf{51.51{±2.10}} &\textbf{50.95{±2.41}} \\
    \bottomrule
    \end{tabular}}
	\label{table:sample_rate}
	\end{minipage}
    \vspace{-7pt}
\end{table*}

\subsection{Main Results}
\noindent\textbf{Results under various vision datasets.} Table~\ref{table:sota_table} shows the personalization performance of different methods under various datasets. We can find that our method achieves state-of-the-art performances in all settings. Meanwhile, we visualize the training curves of the model in Figure~\ref{fig:acc_curves}. We can find that our method can achieve higher accuracy and converge more smoothly and quickly. 
As the number of clients increases, the local data for each client will decrease. In this case, we find that for traditional pFL methods, like FedProto and Ditto, which keep personalized models local, their performances will be significantly affected. For prediction-head-decoupled methods like FedRoD and FedCP, they usually can get better results than FedAvg, but the gains are marginal. This is because, in feature-skew heterogeneity, knowledge mainly comes from the feature extractors, and the dual classifier heads cannot achieve effective knowledge decoupling. While our FediOS uses a dual-feature-extractor decoupling, and in most cases, it reaches state-of-the-art results. The effectiveness of FediOS's feature decoupling is visualized in Figure~\ref{fig:fedios_feature_vis}, where generic features are extracted and generic knowledge is shared. 

\noindent\textbf{Results regarding generalization.} 
Although our method is designed to improve personalization, interestingly, we find that by effectively decoupling and sharing the generic knowledge, the generalization of the global models can also be improved for FediOS. We test the global models after aggregation on a test dataset with all domains/subclasses to indicate the generalization. In Table \ref{table:global_results}, we show the generalization accuracies of different methods. It is found that our FediOS can still achieve a leading performance among the pFL baselines. 

\noindent\textbf{Results under different $N$ with partial client sampling.} 
Scalability is an essential issue for FL methods~\cite{FedProx}, and when the number of clients $N$ increases, due to the communication budget, it is needed to implement partial client sampling instead of full participation. 
In CIFAR-100, we increase the number of clients and experiment with different client sampling rates. In each round of communication, we randomly select a set of clients for model aggregation according to the sampling rate and then send the aggregated model to all clients. In Table \ref{table:sample_rate}, we can find that our method still performs well among various baselines. 

\noindent\textbf{Results under different local epochs $E$.} Larger number of local epochs $E$ results in fewer required communication round $T$, achieving more communication-efficient FL training. We conduct experiments with larger $E$. In Table~\ref{table:local_epoch}, we find that as $E$ increases, the accuracies of most methods decrease, but our FediOS also maintains the best results.

\begin{table}[!t]
    \footnotesize
    \centering
    \caption{\textbf{Results (\%) under different local epochs ($E$).} The dataset is Digits-Five with $N=5$.}
    \vspace{-0.2cm}
    \resizebox{0.48\textwidth}{!}{
    \begin{tabular}{l|c|c|c|c}
    \toprule
     Methods/$E$ &2 &3 &5 &10\\
    \midrule
    FedAvg \cite{FedAvg}  &86.55{±0.13} &86.62{±0.51}&86.49{±0.23} &86.41{±0.19}\\
    FedBN \cite{FedBN}  &86.27{±0.15} &86.83{±0.29}&86.47{±0.19} &86.34{±0.35}\\
    FedCP \cite{FedCP}  &86.96{±0.41} &86.77{±0.17}&86.65{±0.24} &86.59{±0.31}\\
    FedRoD \cite{FedRoD}  &86.52{±0.48}&86.22{±0.16} &86.40{±0.25} &86.14{±0.34}\\
    \midrule
    \rowcolor{gray!20}\textbf{Our FediOS} &\textbf{88.29{±0.16}} &\textbf{88.40{±0.05}} &\textbf{87.69{±0.18}} &\textbf{87.39{±0.45}}\\
    \bottomrule
    \end{tabular}
    }
    \label{table:local_epoch}
    \vspace{-0.5cm}
\end{table}

	

\noindent\textbf{Results regarding more model architectures.} We verify the effectiveness of our FediOS by using more model architectures, i.e., ResNet18~\cite{ResNet18},  EfficientNet~\cite{EfficientNet}, MobileNetV2~\cite{MobileNetV2}, and DenseNet121~\cite{DeseNet121}. Compared with the previously used four-layer CNN model, these neural networks have more complex structures and contain more parameters. As Table~\ref{table:different_model} indicates, our FediOS can improve personalization in various neural network architectures, which verifies the rationale of our decoupling scheme and the effectiveness of the orthogonal projections. At the same time, we find that as the complexity of the network structure increases, the advantage of FedRoD and FedBN diminishes as it has closer results with FedAvg, whereas our method still has notable performance gains.

\begin{figure}[!t]
 \vspace{-0.5cm}
\centering
    \begin{subfigure}[b]{0.23\textwidth}
        \centering
        \includegraphics[width=\linewidth]{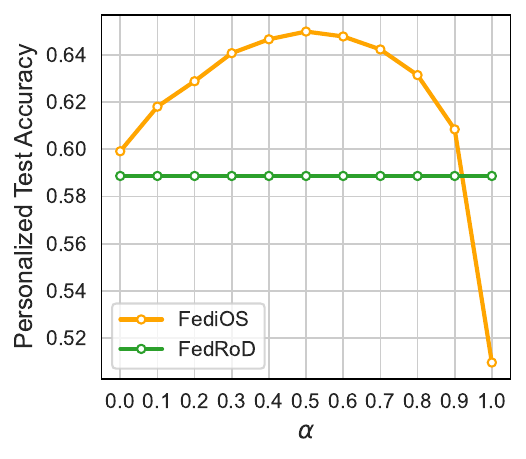}
        \caption{CIFAR-100 with $N=5$.}
    \end{subfigure}
    \begin{subfigure}[b]{0.23\textwidth}
        \centering
        \includegraphics[width=\linewidth]{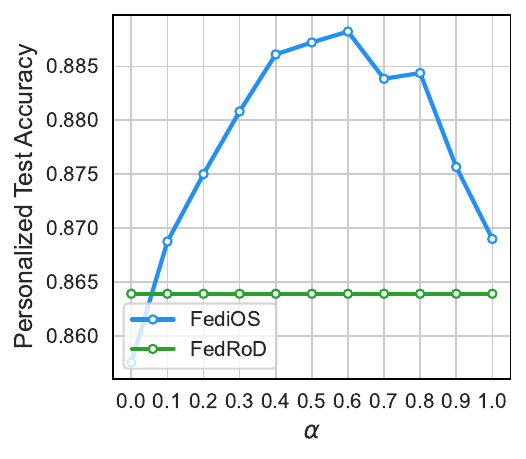}
        \caption{Digits-Five with $N=5$.}
    \end{subfigure}
\caption{\textbf{Results of FediOS under different $\alpha$.} 
}
\label{fig:alpha_acc_curves}
\end{figure}

\begin{table}[t]\footnotesize
\centering
\caption{\textbf{Results (\%) of our FediOS using different $\lambda$.} The dataset is Office-Caltech-10 with $N=4$.}
\vspace{-0.1cm}
\resizebox{0.48\textwidth}{!}{%
\begin{tabular*}{1.05\linewidth}{l|cccc}
    \toprule
    $\lambda$  &\multicolumn{1}{c}{0.1} & \multicolumn{1}{c}{1.0}& \multicolumn{1}{c}{10.0} & \multicolumn{1}{c}{50.0} \\
    \midrule
        Our FediOS&75.98{±1.16} &\textbf{76.07{±0.75}} &75.04{±1.61} &71.87{±0.99} \\
    \bottomrule
  \end{tabular*}
}
 \vspace{-0.2cm}
\label{table:different_lamda}  
\end{table}

\subsection{Sensitivity Analysis of Hyperparameters}
Our FediOS introduces additional hyperparameters $\lambda$ and $\alpha$, and we will present their sensitivity analyses.

\noindent\textbf{Analysis of the hyperparameter $\alpha$.} In FediOS, $\alpha$ controls the proportion of the generic feature when fusing features, and it also reveals the balance between the generic and personalized knowledge. In our implementation, we generally keep $\alpha=0.5$ to keep the uniform balance. While in Figure~\ref{fig:alpha_acc_curves}, we make a sensitivity analysis of $\alpha$ to see the optimal values of $\alpha$ under different datasets. 
Generally, $\alpha$ has a wide range of values to be superior to FedRoD, so FediOS is not sensitive to $\alpha$. 
It has the optimal values around 0.5; for CIFAR-100, the optimal value is 0.5, and for Digits-Five, the value is 0.6. A higher optimal value of $\alpha$ for Digits-Five means that it benefits more on the generic knowledge sharing. The visualization of case samples in each dataset is in the appendix. It is intuitively reasonable that digital numbers share more common and simpler patterns across domains compared with CIFAR-100, therefore, Digits-Five is prone to have higher weights of the generic feature. Intriguingly, it reveals that the intrinsic generic-or-personalized feature structures of the datasets can be explicitly explored by the optimal $\alpha$ of FediOS. 

\noindent\textbf{Analysis of the hyperparameter $\lambda$.} 
In FediOS, the hyperparameter $\lambda$ is for controlling the strength of the orthogonal regularization loss. 
We set $\lambda$ to different values to explore its impact on the algorithm performance. The larger the value of $\lambda$, the more attention the model will pay to the orthogonality of generic features and personalized features. As Table~\ref{table:different_lamda} shows, we observe an optimization-regularization tradeoff regarding different $\lambda$. When $\lambda$ is small and appropriate, it can further strengthen the orthogonality of generic and personalized features, thus, improving personalization. However, if $\lambda$ is too large, the orthogonality regularization will limit the optimization of learning effective representations, and it will even cause negative effects. We suggest to keep $\lambda$ small and around 1 to have a mild and appropriate regularization. We also find that $\lambda$ is not sensitive if being kept small.

\subsection{Ablation Study}
We conduct the ablation study of FediOS's modules in Table~\ref{table:ablation}. It is found that each part of our approach plays an important role, and they work together to achieve better performance. Notably, we notice that orthogonal projection has the most vital impact on the improvement. If without orthogonal projections, the performance shows a huge drop-off, even lower than FedAvg in some cases. This suggests that only using two feature extractors cannot effectively decouple the knowledge, and the orthogonal projections are quite essential. We also find that when FediOS is only without the personalized feature extractor, the model still performs slightly better than FedAvg. It reveals that the projection matrix may map the model parameters into a subspace where the knowledge is more aligned, so it will surpass vanilla FedAvg. Moreover, $\mathcal {L}_{re}$ plays the role of strengthening the orthogonality of generic and personalized features, helping to achieve better results.

\begin{table}[t]\footnotesize
\vspace{-0.5cm}
\centering
\caption{\textbf{Results (\%) of our FediOS using different model architectures.} The dataset is CIFAR-100 with $N=5$.}
\resizebox{0.47\textwidth}{!}{%
\begin{tabular*}{1.12\linewidth}{l|ccc|>{\columncolor{gray!20}}c}
    \toprule
    Model& FedAvg &FedBN & FedRoD& \textbf{FediOS} \\
    \midrule
     ResNet18 \cite{ResNet18}   &64.59{±0.37}  &65.32{±0.46} &65.28{±0.23}  &\textbf{69.21{±0.49}} \\
     EfficientNet \cite{EfficientNet}   &68.45{±0.71} &68.99{±0.58} &69.28{±0.58} &\textbf{73.70{±0.75}} \\
     MobileNetV2 \cite{MobileNetV2}  &73.88{±0.47}  &73.34{±0.33}&73.89{±0.38} &\textbf{76.76{±0.18}} \\
     DenseNet121 \cite{DeseNet121}  &83.79{±0.37}  &83.43{±0.36} &83.75{±0.92}  &\textbf{86.19{±0.46}} \\
    \bottomrule
  \end{tabular*}
}
\label{table:different_model}  
\end{table}

\begin{table}[!t]\footnotesize
\centering
\caption{\textbf{Ablation study of FediOS in terms of personalization.} The datasets are Office-Caltech-10 with $N=4$ and DomainNet with $N=12$.}
\vspace{-0.1cm}
\resizebox{0.48\textwidth}{!}{%
\setlength\tabcolsep{11.55pt}
\begin{tabular*}{1.144\linewidth}{lcc}
    \toprule
    Dataset &Office-Caltech-10 &DomainNet\\
    \midrule
     FedAvg   &69.34{±2.86}  &29.50{±1.22} \\
     \midrule
    Ours w/o Orthogonal Projections &71.77{±0.90}  &26.64{±1.92}\\
    Ours w/o Personalized Extractors  &72.29{±1.00}  &30.48{±1.43} \\
    Ours w/o $\mathcal {L}_{re}$  &75.43{±0.97}  &36.32{±0.86} \\
    \midrule
     \rowcolor{gray!20} \textbf{Ours}    &\textbf{75.98{±1.16}}  &\textbf{37.32{±0.43}}\\
    \bottomrule
  \end{tabular*}
}
 \vspace{-0.2cm}
\label{table:ablation}  
\end{table}

\section{Conclusion}
\label{sec:conclusion}
In this paper, we rethink the architecture decoupling design in feature-skew personalized federated learning and propose FediOS.
Specifically, for each client, We set up a shared generic feature extractor and a local personalized feature extractor. FediOS uses orthogonal projections to map the generic features into one common subspace and scatter the personalized features into different subspaces to achieve feature decoupling. As validated by extensive experiments, our method achieves state-of-the-art feature-skew personalization results compared to strong baselines and can also improve generalization. Furthermore, we gained insights into the effectiveness and applicability of our approach.

{
    \small
    \bibliographystyle{ieeenat_fullname}
    \bibliography{main}
}

\newpage
\appendix
{\LARGE \textbf{Appendix}}
\section{Pseudo Code of FediOS}
In Algorithm \ref{alg:algorithm}, we show the detailed training stage of our approach. In the training process, each client $i$ will simultaneously train a global feature extractor $W^g_i$ and prediction head $H_i$ received from the server, as well as a personalized feature extractor $W^p_i$ that remains local, guided by the loss function $\mathcal{L}$ in Equation (12) of the main paper.

\begin{algorithm}[th]
\caption{FediOS Training Stage}
\label{alg:algorithm}
\textbf{Input}: $N$ clients with local data $\mathcal D $, communication round $T$, local epoch $E$, local learning rate $\eta$, projection layers $\mathbf{P} = \{P^g, P^p_{1},\dots, P^p_{N}\}$;

\textbf{Parameter}: initial generic feature extractor $W^g$, initial personalized feature extractor $W^p$, initial classifier head $H$, and hyperparameters $\alpha, \lambda$;

\textbf{Output}: a final generic feature extractor, $N$ final personalized feature extractors, and a final classifier head;
\begin{algorithmic}[1] 
\STATE Server sends projection layers and initial models to client $i$, $\forall i \in [N]$.
\FOR{each round $t=1,\dots, T$}
        \STATE \texttt{\# Client updates}
        \FOR{each client $i, i\in[N]$ \textbf{in parallel}}
        \STATE set $W^{g,t}_i \leftarrow {W}^{g,t-1}, W^{p,t}_i \leftarrow {W}^{p,t-1}_i, H^{t}_{i} \leftarrow {H}^{t-1}$;
            \FOR{each epoch $e=1,\dots, E$}
                    \FOR{$({x}_{k},{y}_{k}) \in \mathcal D_{i}$}
                        \STATE ${ \phi_{i,k}^{g} \gets ({x}_{k};{W}^{g}_i,{P}^g) }$ in Eq. (8), 
                        \STATE ${ \phi_{i,k}^{p} \gets ({x}_{k};{W}^{p}_i,{P}^p_{i})}$ in Eq. (8),
                        \STATE ${ \phi^f_{i,k} \gets \alpha\phi_{i,k}^{g} + (1-\alpha)\phi_{i,k}^{p}}$;
                        \STATE Compute loss $\mathcal{L}$ by Eq. (12);
                        \STATE ${W}^{g,t}_{i}\gets{W}^{g,t}_{i}-\eta \nabla \mathcal{L}$;
                        \STATE ${W}^{p,t}_{i}\gets{W}^{p,t}_{i}-\eta \nabla \mathcal{L}$;
                        \STATE ${H}^{t}_{i}\gets{H}^{t}_{i}-\eta \nabla \mathcal{L}$;
                    \ENDFOR
            \ENDFOR
        \ENDFOR
        \STATE \texttt{\# Server aggregation}
        \STATE ${W}^{g,t+1} = \sum_{i \in N} \frac{\lvert \mathcal D_{i} \rvert}{\lvert \mathcal D \rvert}{W}^{g,t}_{i}$;
        \STATE ${H}^{t+1} = \sum_{i \in N} \frac{\lvert \mathcal D_{i} \rvert}{\lvert \mathcal D \rvert}{H}_{i}^{t}$;

    \ENDFOR
\STATE \textbf{return} Obtain global model $W^{g,T}, H^T$ and personalized feature extractor  $W^{p,T}_{i}$ for client $i, \forall i \in [N]$.
\end{algorithmic}
\end{algorithm}

\section{Experimental Details}
\subsection{Datasets}
\paragraph{Office-Caltech-10 \cite{office}.} Office-Caltech-10 has four data sources: the Amazon merchant website, the Caltech-101 dataset, a high-resolution DSLR camera, and a webcam. Each source represents a domain, and each domain has the same 10 classes for classification tasks. 

\paragraph{DomainNet \cite{domainnet}.} DomainNet is a dataset containing images from six domains. Each domain represents a style, namely Clipart, Infograph, Painting, Quickdraw, Real, and Sketch. The DomainNet contains 345 object categories, following \cite{FedBN}, we choose the top ten most common classes for our experiments. 

\paragraph{Digits-Five \cite{domainnet}.} Digits-Five is a combination of five digits recognition datasets: MNIST, MNIST-M, Synthetic Digits, SVHN, and USPS. Each dataset represents a style and is divided into ten categories from digits 0 to 9.

\paragraph{CIFAR-100 \cite{cifar100}.} Each image in CIFAR-100 has two labels: superclass and subclass. Each subclass can be included in a superclass. There are 500 training images and 100 testing images per subclass. CIFAR-100 has 20 superclasses and each superclass has 5 subclasses. The images from different subclasses within each superclass share some common features. 
For example, for the superclass of flowers, there are five subclasses, namely, orchid, poppy, rose, sunflower, and tulip. 

\begin{figure*}[th]
 \vspace{-0.5cm}
\centering
    \begin{subfigure}[b]{0.45\textwidth}
        \centering
        \includegraphics[width=\linewidth]{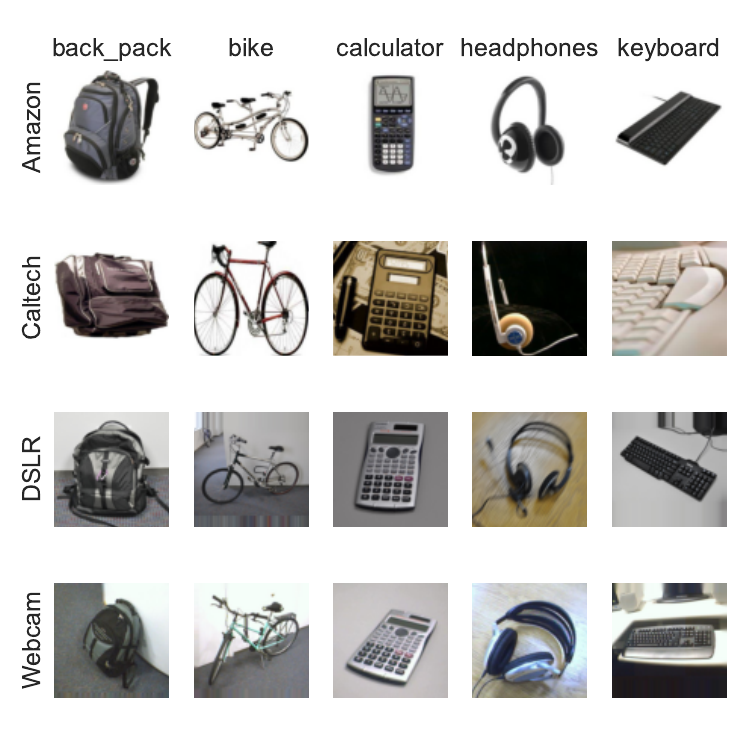}
        \caption{Office-Caltech-10.}
    \end{subfigure}
    \begin{subfigure}[b]{0.45\textwidth}
        \centering
        \includegraphics[width=\linewidth]{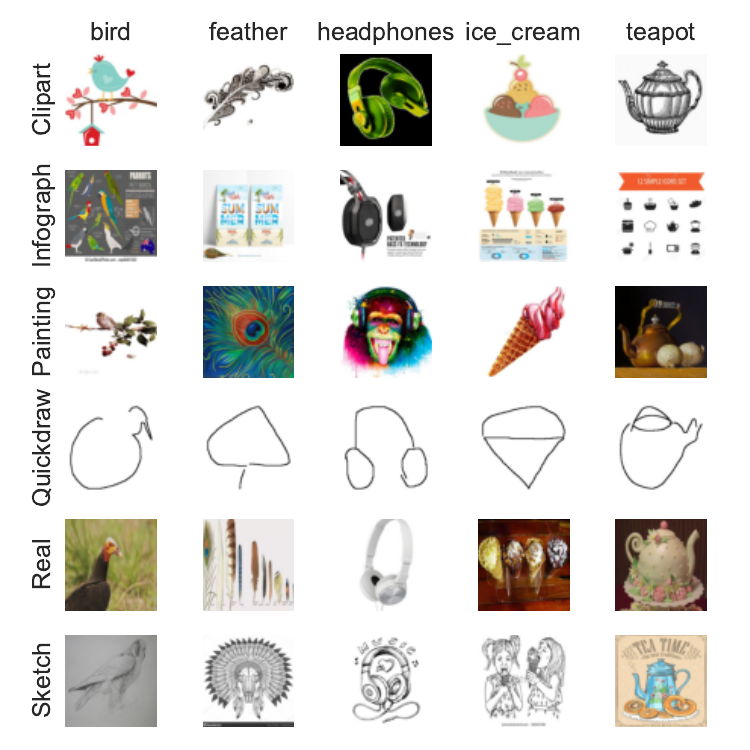}
        \caption{DomainNet.}
    \end{subfigure}
    \begin{subfigure}[b]{0.45\textwidth}
        \centering
        \includegraphics[width=\linewidth]{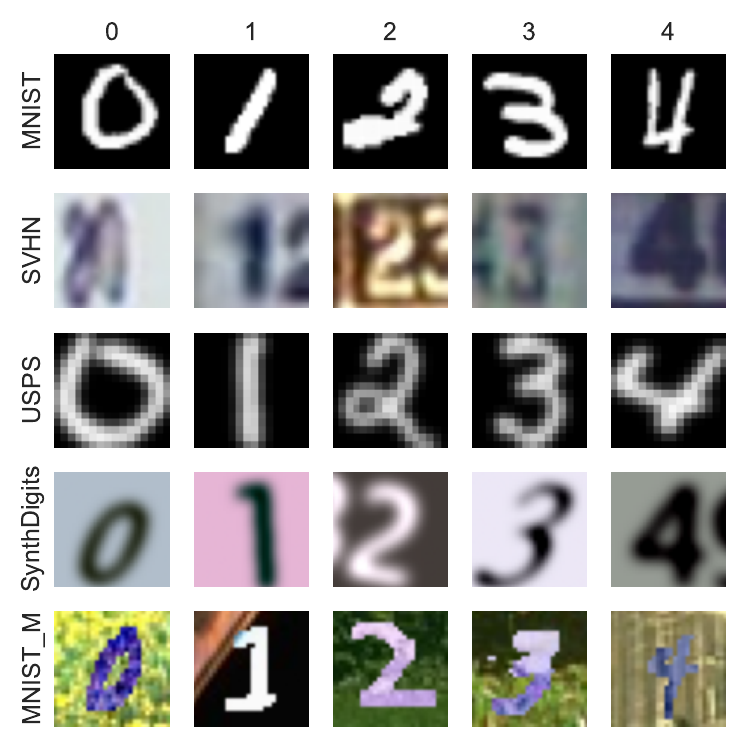}
        \caption{Digits-Five.}
    \end{subfigure}
    \begin{subfigure}[b]{0.45\textwidth}
        \centering
        \includegraphics[width=\linewidth]{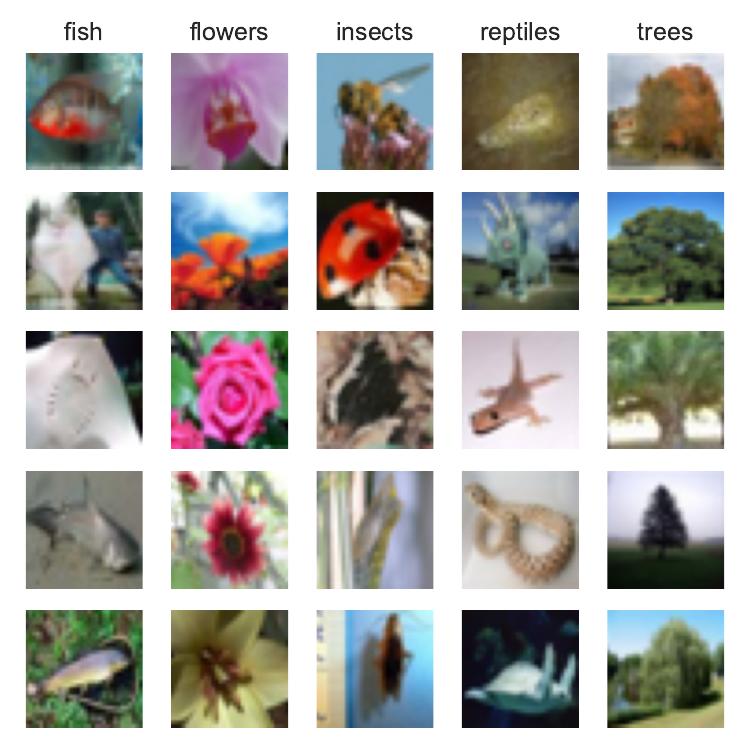}
        \caption{CIFAR-100.}
    \end{subfigure}
    \vspace{-0.3cm}
\caption{\textbf{Demonstration of case samples from four datasets.} For each dataset, we selected five classes for illustration.
}
\label{fig:dataset}
\vspace{-0.2cm}
\end{figure*}

\paragraph{Visualization.} 
To give a more intuitive demonstration of the feature-skew datasets, in Figure \ref{fig:dataset}, we visualize some cases of samples from each dataset.

\subsection{Model Architectures}
For Office-Caltech-10 and DomainNet, we use ResNet-18 and the implementation code refers to ~\cite{FedCP}. For Digits-Five, we use a six-layer CNN that referred from ~\cite{FedBN}. For CIFAR-100, we use a four-layer CNN that referred from ~\cite{FedCP}. We show the details of the network structure in Table \ref{table:six-layer} and Table \ref{table:four-layer}. For the convolutional layer (Conv2D), the parameters we present are input and output dimension, kernel size, stride, and padding. For the max pooling layer (MaxPool2D), we present kernel and stride. For the Batch Normalization layer (BN), we present the channel dimension. For the fully connected layer (FC), we present the input and output dimensions. In the paper, we use the last fully connected layer of the network as the prediction head and the rest as the feature extractor.

\begin{table}[th]\footnotesize
\centering
\caption{\textbf{Model architectures of six-layer CNN~\cite{FedBN} for Digit-Five.} }
\resizebox{0.5\textwidth}{!}{%
\begin{tabular*}{1.05\linewidth}{c|c}
    \toprule
    Layer  &\multicolumn{1}{c}{Details} \\
    
    \midrule
        1 &Conv2D(3, 64, 5, 1, 2), BN(64), ReLU, MaxPool2D(2,2)  \\
    \midrule
        2 &Conv2D(64, 64, 5, 1, 2), BN(64), ReLU, MaxPool2D(2,2)  \\
    \midrule
       3 &Conv2D(64, 128, 5, 1, 2), BN(128), ReLU  \\
    \midrule
    4 &FC(6272,2048), BN(2048), ReLU  \\
    \midrule
    5 &FC(2048,512), BN(512), ReLU  \\
    \midrule
    6 (Head) &FC(512,10)  \\
    
    \bottomrule
  \end{tabular*}
}
 \vspace{-0.2cm}
\label{table:six-layer}  
\end{table}

\begin{table}[th]\footnotesize
\centering
\caption{\textbf{Model architectures of four-layer CNN~\cite{FedCP} for CIFAR-100.} }
\resizebox{0.5\textwidth}{!}{%
\begin{tabular*}{1.05\linewidth}{c|c}
    \toprule
    Layer  &\multicolumn{1}{c}{Details} \\
    
    \midrule
        1 &Conv2D(3, 32, 5, 0, 1), ReLU, MaxPool2D(2,2)  \\
    \midrule
        2 &Conv2D(32, 64, 5, 1, 2), ReLU, MaxPool2D(2,2)  \\
    \midrule
       3 &FC(1600,512), ReLU  \\
    \midrule
    4 (Head) &FC(512,20)\\
    \bottomrule
  \end{tabular*}
}
 \vspace{-0.2cm}
\label{table:four-layer}  
\end{table}

\subsection{Training Details}
\noindent\textbf{Local learning rate and optimizer.} For ResNet-18, the local learning rate (LR) $\eta=0.1$, and for six-layer CNN and four-layer CNN, $\eta=0.01$. For Office-Caltech-10, Digits-Five, and CIFAR-100, we use SGD optimizer with momentum 0.9. For DomainNet, we use SGD optimizer with momentum 0.9 and weight decay $1\times10^{-4}$.

\noindent\textbf{Local epochs $E$ and communication rounds $T$.} If not mentioned otherwise, we use the following settings of $E$ and $T$. For Office-Caltech-10, we set local epochs $E$ = 5, communication rounds $T$ = 120. For DomainNet with the number of clients $N$ = 6, we set local epochs $E$ = 5, communication rounds $T$ = 200. For Digits-Five, we set local epochs $E$ = 1, communication rounds $T$ = 50.  For CIFAR-100 with $N \in \{5, 10, 15\}$, we set local epochs $E$ = 2, communication rounds $T$ = 100. For CIFAR-100 with $N$ = 20, we set local epochs $E$ = 5, communication rounds $T$ = 100.

\noindent\textbf{Hyperparameters.} For our FediOS, we set $\lambda_{\text{FediOS}}=0.1$ in Office-Caltech-10, DomainNet and CIFAR-100, and $\lambda_{\text{FediOS}}=0.0$ in Digits-Five. In this paper, we keep $\alpha_{\text{FediOS}}=0.5$ in all datasets and settings. We set $\alpha_{\text{FedDyn}} = 0.1$ for ResNet-18 and $\alpha_{\text{FedDyn}} = 0.01$ for six-layer CNN and four-layer CNN as suggested in their official implementations or papers. For MOON, we set $\mu_{\text{MOON}}= 1.0$  and $\tau_{\text{MOON}}=0.5$. For Ditto, the learning setting of the personalized model is the same as the one of the global model. For FedProto, we set $\lambda_{\text{FedProto}}=1.0$ as suggested in the paper ~\cite{FedProto}. 
We set $\lambda_{\text{FedCP}}=0.1$ for ResNet-18 and six-layer CNN, and $\lambda_{\text{FedCP}}=5.0$ for four-layer CNN.
For pFedVEM, we train the feature extractor and classifier head of the network with the same learning rate and number of training epochs, and MC sampling is fixed to 5 times as suggested in the paper~\cite{pFedVEM}. 

\end{document}